\documentclass[conference]{IEEEtran}
\usepackage{cite}
\ifCLASSINFOpdf
\else
\fi

\usepackage{graphicx}
\usepackage{multicol}
\usepackage{amsmath}
\usepackage{color}
\usepackage{subfigure}

\newcommand{\xhdr}[1]{\vspace{1.7mm}\noindent{{\bf #1.}}}

\IEEEoverridecommandlockouts

\begin{document}
%
% paper title
% Titles are generally capitalized except for words such as a, an, and, as,
% at, but, by, for, in, nor, of, on, or, the, to and up, which are usually
% not capitalized unless they are the first or last word of the title.
% Linebreaks \\ can be used within to get better formatting as desired.
% Do not put math or special symbols in the title.
\title{Drive2Vec: Multiscale State-Space Embedding of Vehicular Sensor Data}

% author names and affiliations
% use a multiple column layout for up to three different
% affiliations
\author{
    \IEEEauthorblockN{David Hallac\IEEEauthorrefmark{2}\IEEEauthorrefmark{1}\thanks{\IEEEauthorrefmark{1} --- DH, SB, and MC contributed equally to this work.}, Suvrat Bhooshan\IEEEauthorrefmark{2}\IEEEauthorrefmark{1}, Michael Chen\IEEEauthorrefmark{2}\IEEEauthorrefmark{1}, Kacem Abida\IEEEauthorrefmark{3},
    Rok Sosi\v{c}\IEEEauthorrefmark{2}, Jure Leskovec\IEEEauthorrefmark{2}}
    \IEEEauthorblockA{\IEEEauthorrefmark{2}Stanford University
    \\ \small\{hallac, suvrat, mvc, rok, jure\}@stanford.edu}
    \IEEEauthorblockA{\IEEEauthorrefmark{3} Volkswagen Electronics Research Laboratory% Group of America, Inc. %- Electronics Research Laboratory
    \\ \small kacem.abida@vw.com}
}

\maketitle

% As a general rule, do not put math, special symbols or citations
% in the abstract
\begin{abstract}

With automobiles becoming increasingly reliant on sensors to perform various driving tasks, it is important to encode the relevant CAN bus sensor data in a way that captures the general state of the vehicle in a compact form. In this paper, we develop a deep learning-based method, called \emph{Drive2Vec}, for embedding such sensor data in a low-dimensional yet actionable form. Our method is based on stacked gated recurrent units (GRUs). It accepts a short interval of automobile sensor data as input and computes a low-dimensional representation of that data, which can then be used to accurately solve a range of tasks. With this representation, we (1) predict the exact values of the sensors in the short term (up to three seconds in the future), (2) forecast the long-term average values of these same sensors, (3) infer additional contextual information that is not encoded in the data, including the identity of the driver behind the wheel, and (4) build a knowledge base that can be used to auto-label data and identify risky states. We evaluate our approach on a dataset collected by Audi, which equipped a fleet of test vehicles with data loggers to store all sensor readings on 2,098 hours of driving on real roads. We show in several experiments that our method outperforms other baselines by up to 90\%, and we further demonstrate how these embeddings of sensor data can be used to solve a variety of real-world automotive applications.

\end{abstract}

% no keywords

% For peer review papers, you can put extra information on the cover
% page as needed:
% \ifCLASSOPTIONpeerreview
% \begin{center} \bfseries EDICS Category: 3-BBND \end{center}
% \fi
%
% For peerreview papers, this IEEEtran command inserts a page break and
% creates the second title. It will be ignored for other modes.
\IEEEpeerreviewmaketitle

\section{Introduction}

With modern automobiles containing hundreds or even thousands of sensors \cite{F:08}, cars are increasingly relying on sensor data to adapt and react to the vehicle's environment. For example,  this data can be used to automatically adjust to different road conditions \cite{FB:02} or even to stop with the emergency braking when about to hit an object \cite{CEB:10}. Beyond specific applications, there is also value in the raw data itself. This is because the sensor data, which comes from the automobile's Controller Area Network (CAN) bus \cite{LLL:08}, provides a comprehensive picture of the vehicle's \emph{state}, including the particular condition and environment that the car is in at that specific moment in time. However, as more and more sensors are added, it is becoming increasingly difficult to define, evaluate, understand, and eventually analyze the vehicle's state. 

Encoding sensor data to learn the state of the vehicle is valuable because it can be used for situational awareness and to identify specific driving scenarios (i.e., as a knowledge base \cite{MCCD:13}), or even to predict driver actions in the near future, such as turning on a blinker or a lane change. Building this type of knowledge base is especially important in the age of connected, and eventually autonomous, cars. A true state of the car is useful across a variety of contexts, from low-level engine optimization to high-level anticipation and driver/cabin settings, which can improve efficiency, safety, and the driving experience as a whole. When anticipating driver actions before they occur, for example, the vehicle can optimize fuel flow based on how hard it predicts the driver will push the gas pedal in the short-term future \cite{BBLV:18}. Or, vehicle-to-vehicle communication systems could alert neighboring cars when a driver is about to change lanes \cite{SW+:14}. 

However, it is a challenge to aggregate all of this sensor data into a single, compact state. It is unclear how to mathematically define such a state, which sensor data is significant, and how to ensure that the state representation is actionable on a variety of tasks. Additionally, state representations are more powerful when they can be embedded in a compact (low-dimensional) representation. A compact representation is critical in the resource-constrained environment of an automobile, since it is more efficient for storage/transmission and can be transferred easily to any model or decision system. Low-dimensional representations also improve interpretability and reduce overfitting. This need for low-dimensionality adds additional constraints to the already challenging problem of defining and learning a state for automobiles.

In this paper, we develop a method, which we call \emph{Drive2Vec}, of capturing the state of an automobile. Our main contribution is that we propose a novel way of representing the vehicle's state, which we define as a low-dimensional vector that is predictive of both the short and long-term future of the car.
While standard dimensionality reduction techniques \cite{GR:70, HS:06} allow us to understand what the car is currently doing, our definition of state is more nuanced, as it operates on multiple granularities at once.
We require that the state predicts not only what the car is about to do, but also what the car will be doing over the long term. Specifically, in addition to being able to uncover \emph{first-order effects}, or short-term actions (i.e., about to turn right, slow down, etc.), this embedding must also contain information about long-term \emph{second-order effects}, highlighting the environment that the vehicle is in and the characteristics of the driver behind the wheel. For example, the expected long-term average velocity will be very different if the car is in a crowded city compared to an open highway. These environmental factors affect the sensor values over the long-term, and thus can be used to anticipate average values over long time horizons. The advantages of our Drive2Vec method are that the state is compact (low-dimensional), it can be inferred in real-time, it works with both floating point and boolean-valued sensors, and the state estimator can be continually improved as data arrives in a streaming setting.

Drive2Vec uses a deep learning model based on stacked gated recurrent unit (GRU) cells \cite{CMBB:14, CGCB:15}. 
These GRUs are a specific type of recurrent neural network (RNN) cell, capable of learning long-term temporal dependencies in sequential time series data \cite{HS:97}. Our model takes as input a short segment of sensor data and returns a low-dimensional embedding representing the car's state. We analyze and evaluate our method on an anonymized dataset, collected by Audi, where contractors in Ingolstadt, Germany drove modified Audi A3 vehicles equipped with data loggers to store all their sensor readings. Our dataset contains 665 boolean and floating point value sensors and was sampled at 10Hz. We develop a method to encode each sample in just 64 dimensions (i.e. floating point numbers), less than 10\% of the original dataset size.
We do so by taking a short one-second window of time, ending at that timestep, and passing it through our neural network to embed it in a low-dimensional format. Even with this reduction in size, our embedding is able to accurately accomplish a number of tasks. We evaluate Drive2Vec by using it to predict the exact values of all 665 sensors in the short-term (up to three seconds in the future), as well as average sensor values in the long term (up to 100 seconds), outperforming other baselines by up to 90\%. Next, we use these same embeddings to learn contextual information that is not in the sensor data, specifically the identity of the driver. 
We then examine the robustness of our method to various embedding sizes, and we demonstrate the method on three different application case studies. First, we show how Drive2Vec can be used to auto-label common driving actions, such as turns and brake slams. Second, we analyze examples of ``hard braking'', showing that our embeddings are able to identify these risky states even before the actual braking occurs. Finally, we look at the temporal evolutions of these embeddings, focusing on locations where the embedding undergoes a sudden shift. We see that these sudden shifts occur when performing complex short-term maneuvers (i.e., turns) while also changing between different road types, for example from a highway to a rural road, which affects the long-term forecast as well.

Overall, the main contributions of our paper are:
\begin{itemize}
\item We formally define a car's state as a low-dimensional representation that is predictive of both its short and long-term future sensor values.
\item We develop Drive2Vec, a deep learning-based method of learning this state from CAN bus data.
\item We evaluate our method on a large dataset containing thousands of hours of driving data on real roads.
\item We show that our approach is accurately able to solve many useful applications, including driver identification, auto-labeling of data, and more.
\end{itemize}

\xhdr{Related work}
Learning low-dimensional state embeddings is a well-studied problem in various fields, from words in a language \cite{MCCD:13} to nodes in a network \cite{GL:16}. For sensor data, methods for capturing the state have typically relied on approaches that focus on reconstructing the input data, including autoencoders \cite{HS:06} and principal component analysis \cite{WRR:03}. However, these methods are limited in this setting because they are designed to capture only the present (i.e., what the car is doing at that specific moment), but not what the car will do in the near-term and long-term future. With Drive2Vec, in contrast, the same low-dimensional state can be used across many different applications. One such application is short-term prediction, which has typically been solved using Kalman filters \cite{G:11, LSRJ:11}, a specific type of dynamic factor model \cite{M:85}, or using Box-Jenkins ARIMA models \cite{P:09}. However, these approaches are only meant for short-term time series forecasting, and it is difficult to extend them to the various other tasks that we perform. Furthermore, these methods are not well suited for IoT applications due to storage and transmission constraints \cite{AHS:07, BAHT:11}, whereas our low-dimensional embedding is compact, runs in real-time, and can aggregate data from many sensors and across many vehicles to continually improve its state estimator. Similarly, there has been work on driver identification \cite{HS+:16, WPNM:17} and auto-labeling of data \cite{P:02}, but these models are typically built for only one specific purpose. They are unable to transfer across different prediction types and thus struggle to extend into more general knowledge-based tasks. There has also been work on discovering discrete embeddings \cite{HVBL:17}, but such methods can only express a limited number of states, rather than an entire multidimensional spectrum of potential candidates. Within continuous-valued time series predictions, Drive2Vec is the first approach (to the best of our knowledge) that attempts to learn an embedding that is predictive of the future at multiple scales, both in the short and long-term, though the multi-granularity approach has been utilized in other domains \cite{LVRH:16}. Even though embedding the state based on predicted future behaviors is challenging, the many potential applications of such work makes it a necessary problem to solve.

\section{Dataset Description}

The dataset was collected by AUDI AG and Audi Electronics Venture. It contains sensor readings from 10 vehicles, driven by a total of 64 drivers for a cumulative 2,098 hours and covering 110,023 kilometers. The 64 drivers, all paid contractors, drove modified Audi A3's that stored all of the car's sensor readings. To capture and save the data in an economically efficient manner, an offline device was used to store all communication from the car's CAN bus.
The sensors were sampled at various rates, but we synchronized them to 10Hz, the rate at which the majority were sampled.

From this CAN bus dataset, we took every float-valued and boolean sensor with non-zero variance. This comprised a total of 665 signals (110 floats, 555 boolean), each of which were recorded 10 times per second. We note here that we included every float and boolean sensor, so some sensors changed very infrequently (i.e., whether the windshield wipers were on/off), while others changed values at almost every timestep (i.e., the steering wheel position). Additionally, since the sensor readings had different scales and units of measurement (i.e., RPM vs. how hard the gas pedal was pressed), all float-valued sensors were normalized to have zero mean and unit variance.
 
\section{Proposed Method}
\begin{figure}[]
    \centering
    \includegraphics[width=0.99\linewidth]{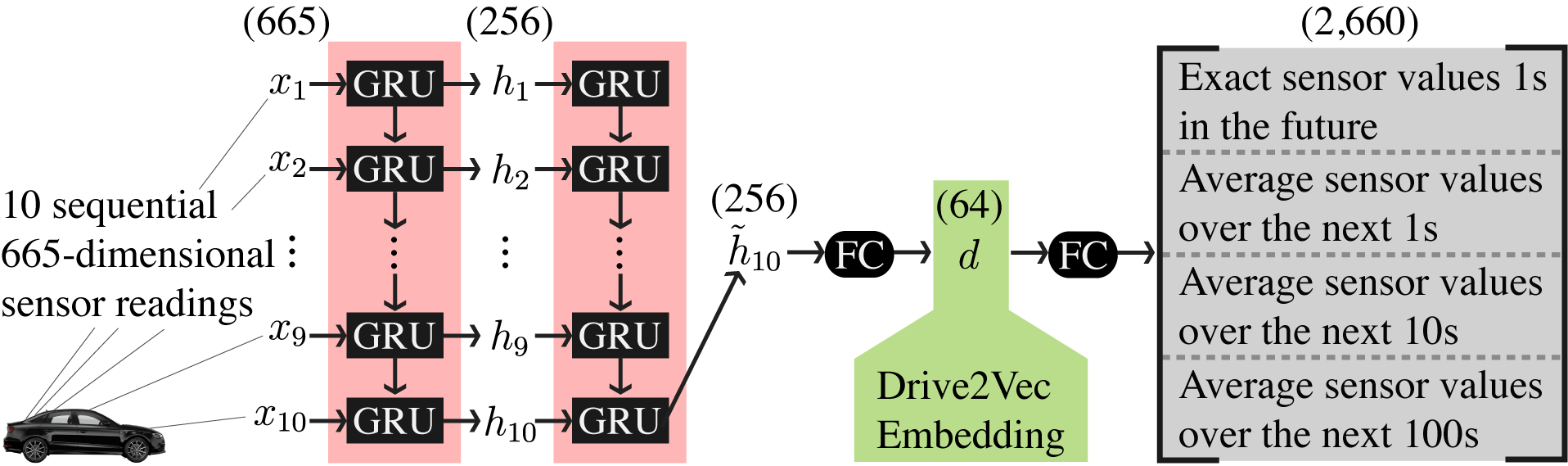}
   \vspace{-7mm}
    \caption{Our model uses two stacked gated recurrent units (GRUs) followed by two fully connected layers (FCs) to convert the 1 second sample containing 6,650 readings down to a 64-dimensional Drive2Vec embedding.}
    \vspace{-5mm}
    \label{fig:GRU}
\end{figure}

Here, we describe our method, which we call \emph{Drive2Vec}, for learning a low-dimensional embedding from a short interval of sensor data. We represent the overall neural network in Figure \ref{fig:GRU}. 
The network accepts 10 timesteps of sensor readings (10 sequential 665-dimensional vectors), which are transformed by subsequent layers to a single 64-dimensional vector (which represents our final Drive2Vec embedding), and finally to a 2,660-dimensional output.
Since the data is sampled at 10Hz, this 10-sample input represents one second of data. These inputs are fed into a gated recurrent unit (GRU) \cite{CGCB:15} of size 256.
 A GRU cell is a type of recurrent neural network (RNN) with an update gate ($z_t$) and reset gate ($r_t$) that update given input $x_t$ and hidden state $h_t$ as
\begin{align*}
z_t = \sigma(W^{(z)} x_t + U^{(z)}h_{t-1}) \\
r_t = \sigma(W^{(r)} x_t + U^{(r)}h_{t-1}),
\end{align*}
where $W$ and $U$ are weight parameters learned by the model.
The hidden state then updates as follows:
\begin{equation}
h_t = z_t \circ h_{t-1} + (1-z_t) \circ \tanh(W^{(h)} x_t + r_t \circ U^{(h)}h_{t-1}).
\label{GRUequation}
\end{equation}
This GRU yields 10 outputs (1 per timestep), each 256-dimensional, which is then fed into a second GRU. We stack these GRU's because this allows different layers of the RNN to operate at different levels of abstraction \cite{PCKB:14}. 
We then send the final output from the last timestep (a size 256 vector) to a series of two dense fully connected layers. The first layer converts the GRU output from size $256 \to 64$ using an exponential linear activation \cite{CUH:15}. The output of this first layer gives us the vehicle's 64-dimensional Drive2Vec embedding, which we show in Section IV to be transferrable across a variety of useful prediction tasks.
The final dense layer goes from  $64 \to 2,660$, where the 2,660-dimensional output can be split into four sections, each of size 665. They are:
\begin{enumerate}
\item \emph{Exact} sensor values of all 665 sensors exactly 1 second after the timestep of the last input. Because our vehicle state involves both floating points and booleans, we optimize the booleans with binary cross-entropy \cite{SJ:80} and floats with mean-squared error loss functions.
\item Average sensor values of all 665 sensors over the next 1 second. Because the average value of booleans is typically a fraction, we use mean-squared error for every signal (including booleans).
\item 665 average sensor values over the next 10 seconds.
\item 665 average sensor values over the next 100 seconds.
\end{enumerate}

All of the above losses are then summed together and optimized in TensorFlow using the Adam optimizer \cite{KB:14}. 
To train our network, we split the automobile data by session, so that overlapping windows are not split between the training, validation, and test set. We use 80\% of the data to train, leaving 10\% each for the validation and test sets.

\section{Experiments}
Drive2Vec embeddings can be used for a variety of tasks. Here, we demonstrate the power and generality of our method by training our 64-dimensional embedding on the Audi data and analyzing its usefulness across several important prediction tasks. It is important to note here that all experiments are performed using the exact same low-dimensional representation (instead of training a new embedding for each experiment), since we aim to show that a single embedding, specifically, the one generated by Drive2Vec, is general enough to solve all these tasks. First we evaluate its ability to predict both the short and long-term future sensor values of the car. We next examine the robustness of our method to the size of the low-dimensional embedding. Then, we show how Drive2Vec can infer the identity of the driver behind the wheel, even though the driver ID is not directly indicated by any of the sensors. Finally, in Section \ref{caseStudies}, we use these embeddings to learn actionable and accurate insights about the state of the car in three real-world case studies. 
Here, we train our model on the training set and use the validation set to optimize our parameters, but all results (in both the experiments and the case studies) are evaluated on a separate hold-out test set.

We compare our GRU-based method with several baselines:
\begin{itemize}
\item Short-only D2V --- similar to Drive2Vec, but instead of the 2,660-dimensional output, only train the network on the \emph{exact} sensor values (665 total) at $t+1$ seconds.
\item Long-only D2V --- similar to Drive2Vec, but only train on the \emph{average} sensor values over the next 100 seconds.
\item PCA 64 --- run principal component analysis (PCA) on the original 665 sensors to reduce to 64-dimensions.
\item Last timestep - repeat the most recent timestep.
\end{itemize}

\xhdr{Predicting the future}
We first use our embeddings to predict the values of all 665 sensors in both the short and long-term future. In Table \ref{shortLongTermError}, we plot the test set error for two tasks: predicting the exact values of the 665 signals exactly 1 second in the future (Short prediction), and predicting the average sensor values over the next 100 seconds (Long prediction). We note that ``Last timestep'' simply predicts a repeat of the most recent 665 sensor values, but Drive2Vec and the other three baselines are all trained the same way. For these methods, we take the 64-dimensional embeddings as input and train a regression model (i.e.\ a single fully-connected layer) to predict the 665 sensor values using the training set data. For the short prediction, we train on the sum of the mean-squared error (MSE) (for floats) and the cross-entropy (for booleans). For the long prediction, since even the booleans are represented as decimals, all signals use MSE. We then evaluate results on the test set and report the mean-squared error of our predictions.

As shown in Table \ref{shortLongTermError}, Drive2Vec outperforms all four baselines. Compared with predicting the last timestep, the worst-performing method, Drive2Vec reduces the error by 90.2\% in the short prediction and 69.6\% in the long prediction. Drive2Vec is unique in that it has very low errors in both tasks. It outperforms Short-only D2V on the short prediction, even though Short-only D2V is explicitly trained for this specific task. Similarly, it is tied with Long-only D2V on the long prediction task. However, those baselines are only trained for their one task, which they perform well on, but they perform poorly on the opposite task (Short-only on the long prediction, and vice versa). Drive2Vec, because it is trained to predict the sensor values at multiple scales, obtains very strong results on both tasks, indicating that it is simultaneously able to encode both short and long-term information in the same embedding.

\begin{table}
	\centering
    \caption{Test set MSE for short (exact sensor values 1 second in the future) and long (average sensor values over next 100 seconds) predictions from different embedding methods.}
    \vspace{-2mm}
\begin{tabular}{ c || c | c }
  & Short prediction MSE & Long prediction MSE \\
  \hline			
  Drive2Vec & 0.020 & 0.021 \\
  Short-only D2V & 0.021 & 0.027 \\
  Long-only D2V & 0.052 & 0.021 \\
  PCA 64 & 0.174 & 0.053 \\
  Last timestep & 0.204 & 0.069 \\
\end{tabular}
\vspace{-4.5mm}
    \label{shortLongTermError}
\end{table}

\begin{figure}[]
    \centering
    \includegraphics[width=0.99\linewidth]{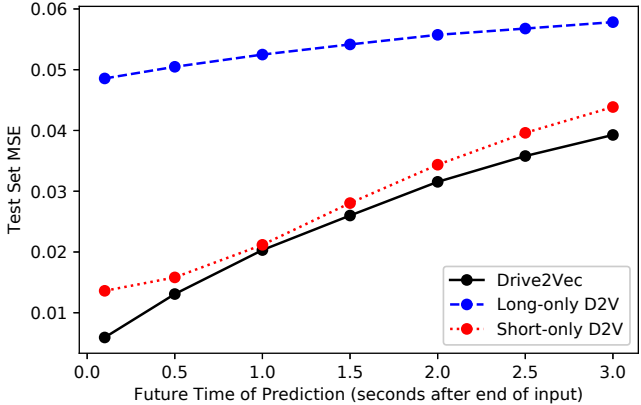}
    \vspace{-8.5mm}
    \caption{Using Drive2Vec's embedding to predict the exact sensor values $K$ seconds in the future. Plot of the MSE vs $K$. As shown, Drive2Vec outperforms short-only D2V, even though the short-only predictor is trained specifically on this short-term prediction task, whereas Drive2Vec is trained on a multiscale approach for predicting both the short and long-term future.}
    \vspace{-6mm}
    \label{fig:shortTerm}
\end{figure}

\xhdr{Short-term predictions}
Next, we look more closely at predicting the state of the car in the near-term future. We train the same short prediction task as before, except here we predict the exact sensor values $K$ seconds in the future, varying $K$ from $0.1$ to $3$. This measures robustness, ensuring that each method does not overfit to predicting exactly one second in the future. For each $K$, we train a new regression model on the training set from our 64-dimensional embedding to predict all 665 sensors. We then evaluate and plot the test set errors in Figure \ref{fig:shortTerm}. For simplicity, we only show the results for the three deep learning-based methods (since PCA and Last timestep perform nearly an order of magnitude worse across all short-term predictions). As shown, the error increases with $K$ for all three methods, meaning they perform better at shorter-term predictions. Long-only D2V performs significantly worse than the other two approaches, since it is only trained on predicting future long-term averages, rather than exact predictions in the short term. Interestingly, Drive2Vec also outperforms the short-only D2V method as well. Here, Drive2Vec and short-only have very similar results at the 1-second offset, but this gap increases as $K$ moves away from 1 second. This implies that short-only D2V may overfit to predicting exactly 1 second in the future, whereas Drive2Vec is more robust to $K$. That is, Drive2Vec is able to leverage insights from training on the long-term future to better predict the sensor values in the short term. This is because Drive2Vec's multiscale approach acts as an implicit regularization on the model, preventing overfitting.

\begin{figure}[]
    \centering
    \includegraphics[width=0.99\linewidth]{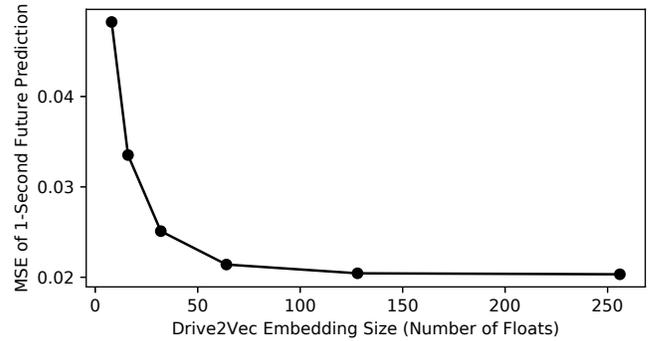}
    \vspace{-8mm}
    \caption{MSE of the short prediction task (exact sensor values 1 second in the future) vs. embedding size of Drive2Vec. Note that the MSE drops dramatically until the embedding is of size 64, after which it flattens out.}
    \vspace{-5mm}
    \label{fig:embeddingSize}
\end{figure}

\xhdr{Effect of the embedding size}
We next evaluate the robustness of Drive2Vec by examining the importance of the embedding size. Recall that our original Drive2Vec method (Figure \ref{fig:GRU}) used 64 floats to encode the data. However, this number could be larger or smaller if needed, though there are tradeoffs in both directions. We train our Drive2Vec model with different embedding sizes and plot the test set short-term prediction error (predicting the exact state 1 second in the future) in Figure \ref{fig:embeddingSize}. As shown, when the Drive2Vec embedding is too small, it does not contain enough information to accurately predict the future of the car. As the embedding size increases, the error goes down, though it begins to flatten out past 64 dimensions. The risk of selecting an embedding that is too large comes from the potential for overfitting, having a less interpretable model, and increased storage and transmission costs. Therefore, even though Drive2Vec could be trained with any embedding size, our 64-dimensional representation occupies a desirable point on the tradeoff curve between accuracy and compactness of our model.

\xhdr{Identifying the driver}
Besides the 665 sensors in the car, Drive2Vec can also be used to infer latent features about the vehicle. One such latent feature is the identify of the driver behind the wheel. We note here that none of the sensors directly indicate the driver identity, but rather that the driver indirectly affects the values of many of the sensors (based on habits such as driving style). We take all 56 drivers who exist in both the training and test sets, and we train a regression model (with one hidden layer of size 32) to predict the driver ID from a single 64-dimensional embedding. This is a 56-way classification problem (with slightly unbalanced classes), and random weighted guessing would be correct only 3.6\% of the time. We then display the micro-$F_1$ score in Table \ref{fig:DriverID}. As shown, Drive2Vec significantly outperforms random guessing, and its micro-$F_1$ score is 32.6\% higher than that of PCA. Our method also slightly outperforms the short-only and long-only D2V baselines. As the score implies, a single 64-dimensional Drive2Vec embedding can correctly identify the driver, out of the set of 56 potential candidates, 51.3\% of the time.

\begin{table}
	\centering
    \caption{Micro-$F_1$ score of driver identification accuracy in 56-way classification task.}
\vspace{-1mm}
\begin{tabular}{ c || c  }
  Method & Micro $F_1$-score \\
  \hline			
  Drive2Vec & 0.513 \\
  Short-only D2V & 0.490 \\
  Long-only D2V & 0.506 \\
  PCA 64 & 0.387 \\
  Random & 0.036
\end{tabular}
\vspace{-6mm}
    \label{fig:DriverID}
\end{table}

\xhdr{Individual sensor predictions}
While all experiments thus far have reported the average mean squared error (MSE) across all 665 signals, this error is not evenly distributed across all of the sensors. In fact, certain sensors are significantly easier to predict than others, and in general, these results are consistent regardless of the embedding method. For example, Drive2Vec has an overall MSE of 0.020 for the short-term prediction task (exact sensor values 1 second in the future, see Table \ref{shortLongTermError}). However, 604 of the 665 sensors had individual MSEs smaller than this value, including gas pedal pressure (0.001), the left/right blinkers (both 0.006), and brake pedal pressure (0.016). On the other hand, many of the sensors with the highest MSEs are related to the instantaneous acceleration of the car, which can vary dramatically from timestep to timestep due to factors such as bumps in the road and are therefore difficult to predict one full second into the future.

\section{Case Studies}\label{caseStudies}
Here, we demonstrate how Drive2Vec can be used to better understand real-world driving data through three case studies.

\begin{figure}[]
    \centering
    \includegraphics[width=0.98\linewidth]{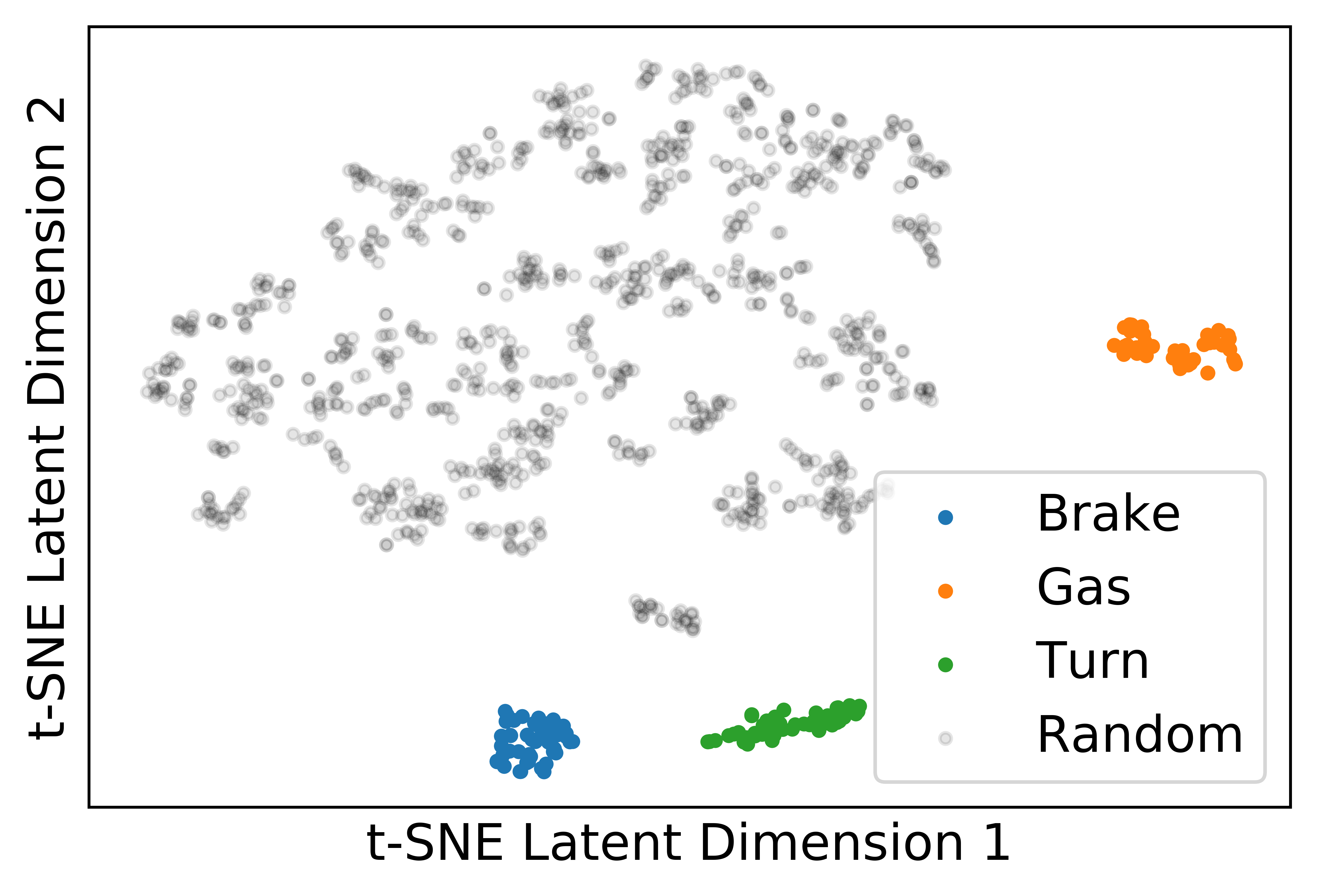}
    \vspace{-3mm}
    \caption{t-SNE plot of brake pedal slams, gas pedal slams, and turns, compared to 1000 random points. The different scenarios have very distinct embeddings.}
    \vspace{-7mm}
    \label{fig:tsne}
\end{figure}

\xhdr{Labeling common actions}
When driving, certain maneuvers can get repeated many times, such as turning, stopping at a red light, or changing lanes. As such, it is important for Drive2Vec to be able to uncover these repeated actions. It is especially valuable if the Drive2Vec embeddings are able to label these actions at the beginning (rather than end) of the maneuver. Here, we take the ``top'' ten brake pedal slams, gas pedal slams, and turns in the test set. These were detected heuristically based on the the maximum difference in brake pedal pressure, gas pedal pressure, and heading, respectively, over a 0.4 second interval. We then take the Drive2Vec embeddings at the beginning of these maneuvers, along with 1000 randomly selected points in the test set, and plot their t-SNE projections \cite{MH:08} in Figure \ref{fig:tsne}. As shown, the three actions have extremely distinct embeddings, and the  different maneuvers are each localized in the plot. This demonstrates that the embeddings of these scenarios have a clear signature that can be used to distinguish it from other points in the dataset. Therefore, the Drive2Vec embeddings can be used to auto-label new data in real time by analyzing the similarity of a streaming embedding with that of several pre-labeled actions, and ``detecting'' a maneuver if the similarity is above a given threshold.

\xhdr{Discovering risky states before they occur}
In addition to labeling data, Drive2Vec can also be used to discover risky states before they actually occur. Here, we focus on the brake pedal slam described above (a difference in brake pedal pressure above a certain threshold in a 0.4 second interval). These are often the most dangerous moments for the driver, as the car needs to quickly slow down to avoid an accident. While there has been much research specifically aimed at emergency braking \cite{CEB:10}, our general Drive2Vec embedding can uncover these risky states as well. We evaluate this hypothesis by taking every time in the test set that the brake difference was above a threshold of  $\epsilon = 25$, which yields 122 ``hard brake'' examples. Once again, we take the beginning of the interval, so this embedding occurs \emph{before} the driver slams on the brake pedal. We took 80\% of these embeddings as our training set, leaving the remaining 20\% as the ``positive'' examples. The ``negative'' examples were the other ~8.5 million points in the original Drive2Vec test set. We then computed a nearest-neighbor cosine similarity score with the training set for both the positive and negative examples. To evaluate our results, we then computed the area under the receiver operating characteristic (AUROC) score, achieving an AUROC of 0.999983. This score shows a near-perfect separation between the positive and negative examples, meaning that almost every ``hard brake'' achieves a higher similarity score than every other point out of the 8.5 million negative examples. Furthermore, the negative examples with the highest similarity scores were all examples of ``hard brakes'' that ended up just below our threshold of $\epsilon = 25$. Overall, the very high AUROC score in this example indicates that Drive2Vec is able to identify this risky state even before the driver begins to slam the brake pedal.

%\begin{figure}[]
%    \centering
%    \includegraphics[width=0.99\linewidth]{FIG/auc_brake.png}
%    \caption{AUROC curve of hard braking. Drive2Vec is able to anticipate these scenarios even before the driver begins to press the brake pedal with very high accuracy.}
%    \label{fig:AUCplot}
%\end{figure}

\begin{figure}[]%!t?
\centering 
  \subfigure[]{\label{fig:turn1}\includegraphics[width=0.485\linewidth]{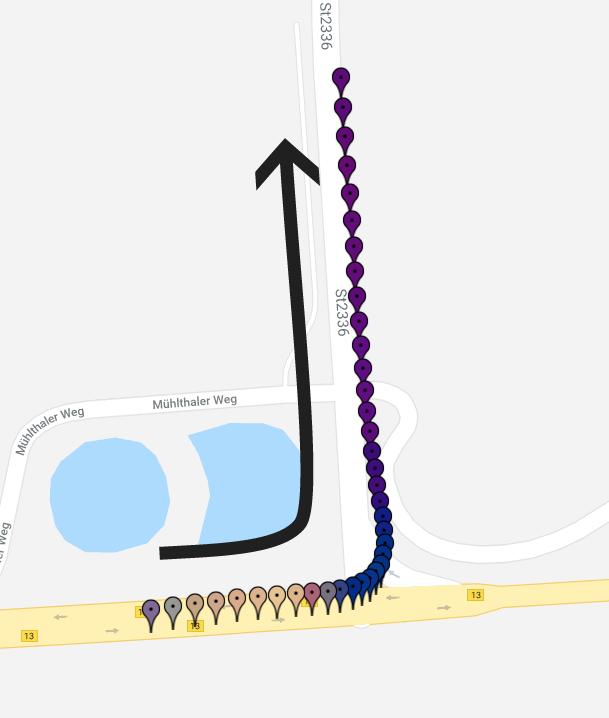}}
  \subfigure[]{\label{fig:turn2}\includegraphics[width=0.5\linewidth]{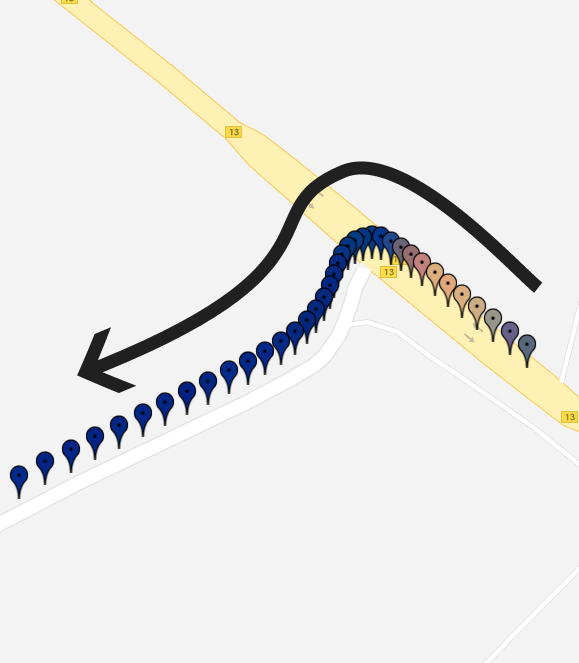}}
  \vspace{-4mm}
   \caption{Two examples of turning from a highway to a rural road. In both cases, the embedding shifts as the turn is made, since both the short and long-term averages of the different signals change dramatically after this turn. }
   \vspace{-5mm}
   \label{fig:turns}
\end{figure}

\xhdr{Temporal evolution of embeddings}
An additional benefit of Drive2Vec is that it can be used to visualize and better understand a driving session. That is, as a vehicle drives along the road, the embeddings form a ``signature'' of the state of the car at that moment in time. In general, these embeddings vary smoothly across time. There are slight shifts, for example, as a car stops at a light or turns at an intersection, but in general, these embeddings only vary slightly across adjacent timesteps. 
However, occasionally, there are times where the embedding undergoes a large shift in a very short interval. These occur when the short and long-term future of the car undergoes a significant shift in expectation. We can analyze when these events occur by plotting the latitude/longitude with an RGB marker, colored by the corresponding 64-dimensional embedding (using PCA to reduce it to 3 dimensions). We notice that this typically occurs when the car undergoes a significant environmental shift, as seen in Figure \ref{fig:turns} where we plot two instances of these sharp breakpoints. We see that in both these examples, the car is turning from a major highway onto a rural road. As the car makes this turn, the short-term predictions of the sensor values changes, the sensor values are very different in turns compared to straightaways. However, as the car straightens out onto the rural road, the long-term expected values of these signals changes as well (slower expected velocity, more expected red lights, etc.). This causes the Drive2Vec embeddings to exhibit a dramatic shift, in this case from beige to purple in the RGB markers.

\section{Conclusion and Future Work}
In this paper, we have developed Drive2Vec, a method for embedding automobile sensor values in a low-dimensional representation that is able to predict the short and long-term future of the car. While experiments show that our method is a very powerful tool, there are several potential directions for further exploration. We leave for future work the analysis of our method to different IoT-connected machines that generate sensor data (i.e., airplanes or ships). Our work could also be extended to learn different types of embeddings, beyond a representation that simply predicts future sensor values. For example, encoding based on ``risk profile'' of component failures would allow this work to be used for predictive maintenance. Overall, the promising results and various extensions highlight the practical applicability of Drive2Vec.

% conference papers do not normally have an appendix

% trigger a \newpage just before the given reference
% number - used to balance the columns on the last page
% adjust value as needed - may need to be readjusted if
% the document is modified later
%\IEEEtriggeratref{8}
% The "triggered" command can be changed if desired:
%\IEEEtriggercmd{\enlargethispage{-5in}}

% references section

% can use a bibliography generated by BibTeX as a .bbl file
% BibTeX documentation can be easily obtained at:
% http://mirror.ctan.org/biblio/bibtex/contrib/doc/
% The IEEEtran BibTeX style support page is at:
% http://www.michaelshell.org/tex/ieeetran/bibtex/
%\bibliographystyle{IEEEtran}
% argument is your BibTeX string definitions and bibliography database(s)
%\bibliography{IEEEabrv,../bib/paper}
%
% <OR> manually copy in the resultant .bbl file
% set second argument of \begin to the number of references
% (used to reserve space for the reference number labels box)

% that's all folks
\end{document}